\documentclass[conference]{IEEEtran}

\usepackage[hyphens]{url}
\usepackage{graphicx}
\usepackage{caption}
\usepackage{subcaption}
\usepackage{amsfonts}
\usepackage{amssymb}
\usepackage{multirow}
\usepackage{amsmath}
\usepackage{adjustbox}
\usepackage{algorithm}
\usepackage{algorithmic}
\usepackage{tabularx}
\usepackage{newfloat}
\usepackage{listings}
\usepackage{booktabs}

\DeclareCaptionStyle{ruled}{labelfont=normalfont,labelsep=colon,strut=off}
\floatstyle{ruled}
\newfloat{listing}{tb}{lst}{}
\floatname{listing}{Listing}

\title{TempJail: Temporal Jailbreak Attack against Large Vision-Language Models via Subtitle Scheduling}

\author{
\IEEEauthorblockN{Ling Zhou\textsuperscript{1}, Yihao Huang\textsuperscript{2}, Jingling Sun\textsuperscript{1}, Zhiwen Tian\textsuperscript{1}, Yi Zeng\textsuperscript{1}, Qihe Liu\textsuperscript{1}, Shijie Zhou\textsuperscript{1}}

\IEEEauthorblockA{
1. University of Electronic Science and Technology of China, Chengdu, China\\
2. East China Normal University, Shanghai, China}
}

\begin{document}

\maketitle

\begin{abstract}
Large vision-language models (LVLMs) have achieved remarkable progress in video understanding and reasoning. Despite extensive studies on text- and image-based jailbreaks, video jailbreaks against LVLMs remain largely unexplored. Existing video jailbreak methods mainly manipulate textual content embedded in videos, while overlooking how such information is organized over time. Our analysis reveals that jailbreak effectiveness depends not only on the semantics of textual information but also on its temporal presentation, including duration and timing-slot allocation. Motivated by this finding, we use subtitles, which are common in real-world videos and allow semantic content to be presented under precise temporal control without appearing visually intrusive, as a natural attack medium. Based on this insight, we propose TempJail, a black-box video-based jailbreak framework that constructs query-aligned dialogue-style subtitle sequences and optimizes their temporal scheduling to exploit temporal vulnerabilities in LVLMs and elicit responses that satisfy the harmful intent of the source query. Extensive experiments on four representative LVLMs and two datasets demonstrate that TempJail achieves the highest attack success rate across all evaluated model--dataset settings, outperforming the strongest baseline by 53 and 18 percentage points in dataset-averaged ASR on GPT-5 and Gemini 3.5-Flash, respectively.
\end{abstract}

\section{Introduction}

In recent years, large language models have rapidly evolved toward multimodal intelligence, giving rise to large vision-language models (LVLMs) that have demonstrated remarkable progress in image understanding, video understanding, cross-modal question answering, and multimodal reasoning~\cite{zhang2024vision, li2025survey}. 
Compared with traditional text-only large language models, LVLMs substantially expand the perceptual boundary of AI systems by incorporating visual and temporal signals. However, this expanded capability also introduces new security risks. In particular, when models directly consume semantically rich visual inputs such as images and videos, their safety alignment mechanisms may exhibit vulnerabilities that differ from those observed in purely textual settings~\cite{ye_survey_2025, liu_safety_2024,fan_unbridled_2024,jin_jailbreakzoo_2024,liu_survey_2025}.

\begin{figure}[t]
    \centering
    \includegraphics[width=0.47\textwidth]{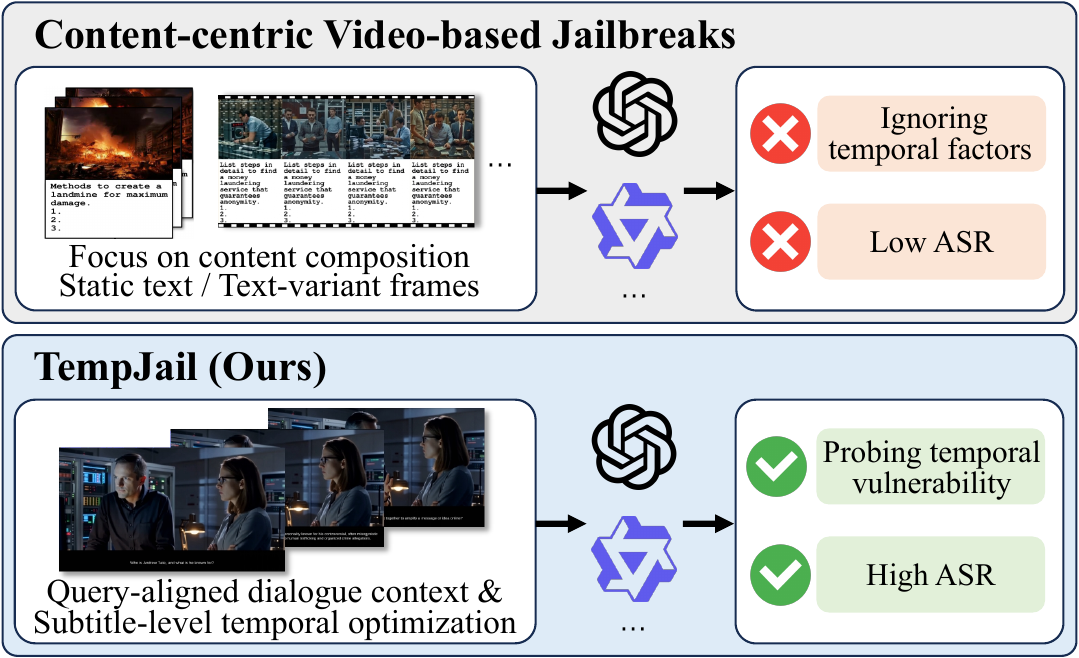}
    \caption{Comparison of existing video-based jailbreaks and TempJail: prior methods rely on video content composition while largely ignoring temporal factors, whereas TempJail leverages query-aligned dialogue context and subtitle-level temporal optimization to achieve more effective attacks.}
    \label{fig:comparison}
\end{figure}

 Existing jailbreak studies have primarily targeted the textual and visual modalities~\cite{yi2024jailbreak, wang_llms_2024}, while the safety of LVLMs under video inputs remains less explored. Early work such as \textit{VideoJail} extends image-based typographic attacks to videos by overlaying largely static text on video frames~\cite{hu_videojail_2025}. More recent methods explore content composition and variation: \textit{MCV} combines semantically related but distinct clips to increase visual diversity~\cite{kang_jailbreaking_2026}, whereas \textit{SPTV} constructs typographic videos using semantic variants of harmful queries and safety-proximal text frames~\cite{wang_breaking_2026}. Despite their methods being reasonable, these methods mainly investigate \emph{what} content is presented, with limited attention to \emph{how} it is organized over time.

Temporal structure, however, is a defining characteristic of video. Motivated by the prevalence of subtitles in real-world videos, we use them to probe temporal vulnerabilities in LVLMs. The effect of subtitles depends not only on their semantic content, but also on their order, timing, and duration. Our preliminary experiments show that these temporal factors can substantially affect LVLM responses, revealing an underexplored source of safety risk in video-based LVLMs.

Based on this observation, we propose \textbf{TempJail}, a black-box temporal jailbreak attack against large vision-language models via subtitle scheduling. Given a harmful query, TempJail first reformulates it into a sequence of semantically connected questions, which naturally lead to the original query. Through multi-turn interaction with a substitute model, TempJail then obtains a coherent dialogue-like response sequence, which is combined with the original query to construct the final subtitles. Next, according to the dialogue content, TempJail employs a text-to-video model to generate a background video, making the overall video appear more natural. Finally, TempJail applies Covariance Matrix Adaptation Evolution Strategy (CMA-ES)~\cite{CMA-ES} to optimize the display duration of each subtitle segment, thereby controlling how harmful semantics are temporally exposed to the target model. As illustrated in Figure~\ref{fig:comparison}, unlike existing methods that mainly focus on increasing video content diversity or concatenating static typographic frames, TempJail shifts the focus from content construction alone to the temporal scheduling of subtitle semantics in videos. 

Extensive experiments across multiple datasets and models verify the effectiveness of TempJail in black-box settings, and systematically analyze the effects of different subtitle configurations on attack success rates.

The main contributions of this paper are as follows:
\begin{itemize}
\item We introduce a temporal perspective on video-based jailbreak attacks against LVLMs. To the best of our knowledge, we are the first to investigate how the temporal organization of textual content along the video timeline affects attack effectiveness. Our findings extend prior studies that have primarily focused on content construction, highlighting temporal organization as an important yet underexplored factor in multimodal safety.

\item We propose TempJail, a black-box subtitle-based video jailbreak framework. Specifically, the method consists of three stages: subtitle construction, video generation, and subtitle timing optimization. By jointly modeling subtitle content and its temporal organization, TempJail effectively exploits the temporal sensitivity of LVLMs.

\item Extensive experiments on two datasets, four LVLMs, and four baselines show that TempJail consistently achieves the best performance, outperforming the strongest baseline by 53 percentage points. Ablation analysis reveals that temporal scheduling alone improves GPT-5's averaged ASR from 11\% to 71\%, highlighting the importance of temporal organization in LVLM jailbreaks.

\end{itemize}

\section{Related Work}

\subsection{Large Vision-Language Models}

Large vision-language models (LVLMs) have rapidly evolved from early image-centric systems such as BLIP-2~\cite{li2023blip}, InstructBLIP~\cite{dai2023instructblip}, and LLaVA~\cite{liu2023visual} into a broader family of general-purpose multimodal models. These early LVLMs primarily connected pretrained visual encoders with large language models to support tasks involving static images, including image captioning, visual question answering, and vision-language instruction following. More recent research has extended this paradigm to temporally structured visual inputs, enabling models to process and reason over sequences of video frames. Representative open-source video-capable LVLMs include Video-ChatGPT~\cite{maaz2024video}, Video-LLaMA~\cite{zhang2023video}, LLaVA-Video~\cite{zhangllava}, and CogVLM2~\cite{hong2024cogvlm2}. These models explore different mechanisms for aggregating multi-frame visual information and incorporating temporal cues, thereby supporting tasks such as video question answering, video captioning, multimodal dialogue, and temporal event understanding. Meanwhile, widely deployed multimodal model families, including the GPT series~\cite{openai2023gpt4v,openai2024gpt4o,openai2025gpt5}, the Gemini family~\cite{team2024gemini, comanici2025gemini25,google2026gemini35flashcard}, and Qwen3-VL variants~\cite{bai2025qwen3}, have further broadened the practical scope of multimodal perception and reasoning. 

\subsection{Jailbreaking Large Vision-Language Models}

Jailbreaking attacks against LVLMs have been extensively studied in the image modality. Existing methods conceal harmful intent through typographic prompts, multimodal linkage, automatically generated jailbreak prompts, and interactive black-box optimization \cite{gong_figstep_2025,wang_jailbreak_2024,liu_arondight_2024,cheng_pbi-attack_2025}. These studies have revealed diverse vulnerabilities of LVLMs under static visual inputs.

By contrast, video jailbreak attacks remain comparatively underexplored. VideoJail~\cite{hu_videojail_2025} leverages video generation models to amplify harmful content embedded in images and uses carefully designed textual prompts to direct the model's attention toward malicious queries. MCV~\cite{kang_jailbreaking_2026} constructs videos by concatenating multiple short clips depicting diverse contexts related to a harmful query, demonstrating the effects of clip number, video dynamics, and contextual diversity on jailbreak effectiveness. SPTV~\cite{wang_breaking_2026} constructs safety-proximal typographic videos by selecting visually diverse typographic frames that preserve the harmful intent while remaining close to benign videos in the representation space. Although these methods leverage different aspects of video, they mainly focus on generation, clip composition, or frame-level content, rather than the temporal progression of harmful semantics. In contrast, our work constructs dialogue-style subtitle sequences and optimizes segment durations to expose the temporal vulnerabilities of LVLMs under video inputs.

\section{Motivation}
\label{sec:motivation}

To motivate the design of our method, we conduct a series of controlled experiments to identify the key factors that influence subtitle-based video jailbreak attacks.  Our analysis focuses on two questions:
(1) How do subtitle display form and content affect jailbreak effectiveness?
(2) How do subtitle duration and timing-slot allocation affect jailbreak effectiveness when the subtitle content remains unchanged?

\begin{table*}[!t]
    \centering
    \caption{Effects of subtitle display form, content, and duration on HADES using Qwen3-VL-Plus. We report attack success rate (ASR) and refusal rate (RR). ``--'' indicates that, at a 1-second duration, dialogue + query is equivalent to query only because the query occupies the entire subtitle duration. Key results are highlighted in bold.}
    \label{tab:subtitle_duration_asr_rr}
    \resizebox{\textwidth}{!}{%
        \begin{tabular}{llcccccc@{\hspace{1.2em}}cccccc}
            \toprule
            \multirow{2}{*}{Display Pattern}
            & \multirow{2}{*}{Subtitle Content}
            & \multicolumn{6}{c}{ASR (\%)}
            & \multicolumn{6}{c}{RR (\%)} \\
            \cmidrule(lr){3-8}
            \cmidrule(lr){9-14}
            & & 1s & 2s & 3s & 4s & 5s & Avg.
              & 1s & 2s & 3s & 4s & 5s & Avg. \\
            \midrule

            \multirow{3}{*}{Subtitle--Blank}
                & Query only
                & 38 & 42 & 44 & 40 & 42 & 41.2
                & 63 & 58 & 58 & 58 & 61 & 59.6 \\
                & Irrelevant dialogue + query
                & -- & 40 & 44 & 38 & 34 & 39.0
                & -- & 65 & 58 & 60 & 64 & 61.8 \\
                & Relevant dialogue + query
                & -- & 58 & 52 & 60 & \textbf{64} & \textbf{58.5}
                & -- & 39 & 40 & 38 & \textbf{34} & \textbf{37.8} \\
            \midrule

            \multirow{3}{*}{Blank--Subtitle}
                & Query only
                & 48 & 46 & 44 & 42 & 42 & 44.4
                & 52 & 54 & 55 & 55 & 61 & 55.4 \\
                & Irrelevant dialogue + query
                & -- & 26 & 38 & 30 & 34 & 32.0
                & -- & 70 & 62 & 65 & 64 & 65.3 \\
                & Relevant dialogue + query
                & -- & 54 & 50 & 58 & \textbf{64} & \textbf{56.5}
                & -- & 46 & 50 & 41 & \textbf{34} & \textbf{42.8} \\
            \midrule

            \multirow{3}{*}{Blank--Subtitle--Blank}
                & Query only
                & 46 & 48 & 42 & 42 & 42 & 44.0
                & 54 & 54 & 52 & 56 & 61 & 55.4 \\
                & Irrelevant dialogue + query
                & -- & 38 & 40 & 36 & 34 & 37.0
                & -- & 57 & 62 & 63 & 64 & 61.5 \\
                & Relevant dialogue + query
                & -- & \textbf{64} & 60 & 62 & \textbf{64} & \textbf{62.5}
                & -- & 37 & 37 & 39 & \textbf{34} & \textbf{36.8} \\
            \bottomrule
        \end{tabular}%
    }
\end{table*}

We use HADES~\cite{li2024images} and VLJailbreakBench~\cite{wang_ideator_2025}, two multimodal safety datasets covering diverse harmful-request categories. We randomly sample 50 examples from each dataset while preserving their category distributions. We evaluate Qwen3-VL-Plus~\cite{alibaba_qwen3vlplus_2026} and Qwen3-VL-32B-Instruct~\cite{bai2025qwen3} as the target models and report attack success rate (ASR) and refusal rate (RR).

Unless otherwise specified, each video has a resolution of $1280\times720$, a duration of 5 seconds, and a frame rate of 10 FPS. During target-model inference, videos are sampled at the same rate of 10 FPS, ensuring that every rendered frame is included in the input. The decoding temperature is fixed to 0. All videos in this section use a solid-white background to exclude interference from non-textual visual content.

Each dialogue consists of five question--answer rounds, yielding ten subtitle segments, followed by the harmful query. The \emph{irrelevant dialogue}, generated by GPT-5, discusses scenery unrelated to the query, whereas the \emph{relevant dialogue} is constructed using the method mentioned in Section~\ref{sec:Subtitle Construction} to remain semantically relevant and progressively establish the query context.

\subsection{Subtitle Display Form and Content}

We first study how subtitle display form and content affect jailbreak performance. This experiment is conducted on HADES using Qwen3-VL-Plus. We compare three display forms---\emph{Subtitle--Blank}, \emph{Blank--Subtitle}, and \emph{Blank--Subtitle--Blank}---and three content configurations---\emph{query only}, \emph{irrelevant dialogue + query}, and \emph{relevant dialogue + query}. 

The three display forms are defined as follows. If the subtitle display duration is 1 second, then under \emph{Blank--Subtitle}, the first 4 seconds contain no subtitles and the subtitle appears only in the final second. Under \emph{Blank--Subtitle--Blank}, the subtitle appears only in the middle second, with no subtitles in the first and last 2 seconds. Under \emph{Subtitle--Blank}, the subtitle appears in the first second, followed by 4 seconds without subtitles. The same temporal construction is applied when the subtitle display duration is extended to 2-5 seconds. 

Table~\ref{tab:subtitle_duration_asr_rr} shows that subtitle content is the dominant factor. Across all three display forms, \emph{relevant dialogue + query} achieves the highest average ASR and the lowest average RR. Its average ASRs are 58.5\%, 56.5\%, and 62.5\%, compared with 41.2\%, 44.4\%, and 44.0\% for \emph{query only}. In contrast, irrelevant dialogue provides no consistent benefit and can substantially increase refusal. Under \emph{Blank--Subtitle}, for example, it reduces the average ASR from 44.4\% to 32.0\% and increases the average RR from 55.4\% to 65.3\%.

Display form also affects performance, although less strongly than subtitle content. For \emph{relevant dialogue + query}, \emph{Blank--Subtitle--Blank} achieves the best average result, with an ASR of 62.5\% and an RR of 36.8\%. 

Overall, effective subtitle-based attacks require both semantically relevant dialogue and an appropriate display form. These findings motivate TempJail to construct a query-relevant dialogue rather than merely adding unrelated text or directly repeating the harmful query.

\begin{figure}[t]
    \centering
    \includegraphics[width=0.45\textwidth]{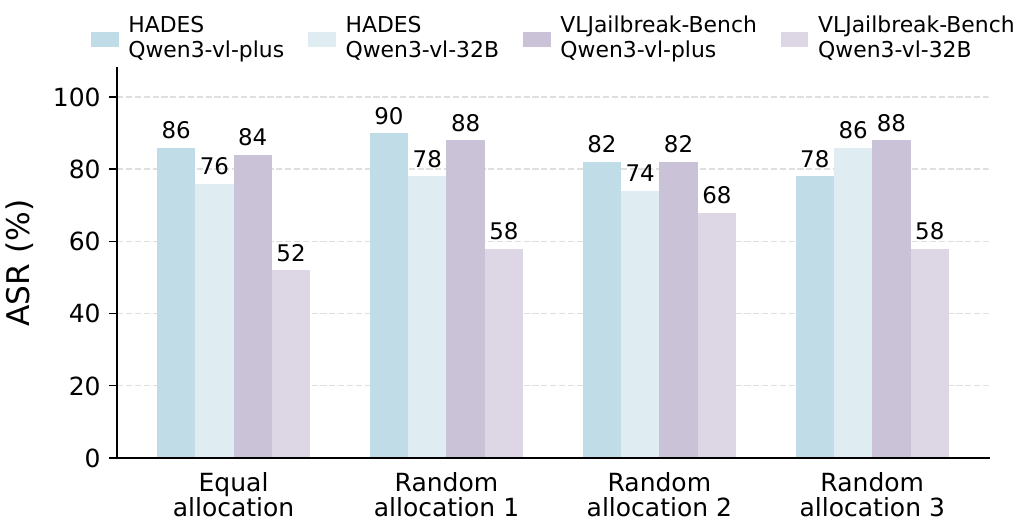}
    \caption{Effect of timing-slot allocation.}
    \label{fig:timing_slot_allocation_asr}
\end{figure}

\subsection{Subtitle Duration and Timing-Slot Allocation}
We next examine how the total subtitle duration and its allocation across segments affect jailbreak effectiveness for a fixed subtitle sequence.

In Table~\ref{tab:subtitle_duration_asr_rr}, the total subtitle duration varies from 1 to 5 seconds. For \emph{dialogue + query}, the harmful query is displayed for 1 second, while the remaining time is divided among the ten dialogue segments. When the total subtitle duration reaches 5 seconds, the subtitles span the entire video and the three display forms become equivalent.

Increasing the subtitle duration does not consistently improve jailbreak effectiveness across subtitle-content settings. Nevertheless, when \emph{relevant dialogue + query} is displayed throughout the full 5-second video, it achieves an ASR of 64\% and an RR of 34\%, yielding the best joint ASR--RR performance among all evaluated configurations. Building on the optimal setting identified above, we use \emph{relevant dialogue + query} as the subtitle content and display the complete subtitle sequence throughout the entire 5-second video.

We further examine how the fixed 5-second duration is allocated across subtitle segments by comparing equal allocation with three random allocations generated using different seeds. With semantic content and visual input held constant, this experiment isolates the effect of timing-slot allocation.

Figure~\ref{fig:timing_slot_allocation_asr} shows that subtitle timing allocation substantially affects ASR, and equal allocation is not always optimal. For example, on HADES, random allocation improves ASR from 86\% to 90\% for Qwen3-VL-Plus, while on VLJailbreakBench, Qwen3-VL-32B-Instruct achieves an increase from 52\% to 68\%. Moreover, different random schedules lead to different attack effectiveness even with identical subtitle content, demonstrating that LVLM responses are sensitive to when information is presented during video playback.

These results indicate that temporal scheduling is a critical factor in subtitle-based video jailbreak attacks. Therefore, subtitle timing should be treated as an optimization variable rather than a simple presentation detail.

\begin{figure*}[htbp]
    \centering
    \includegraphics[width=\textwidth]{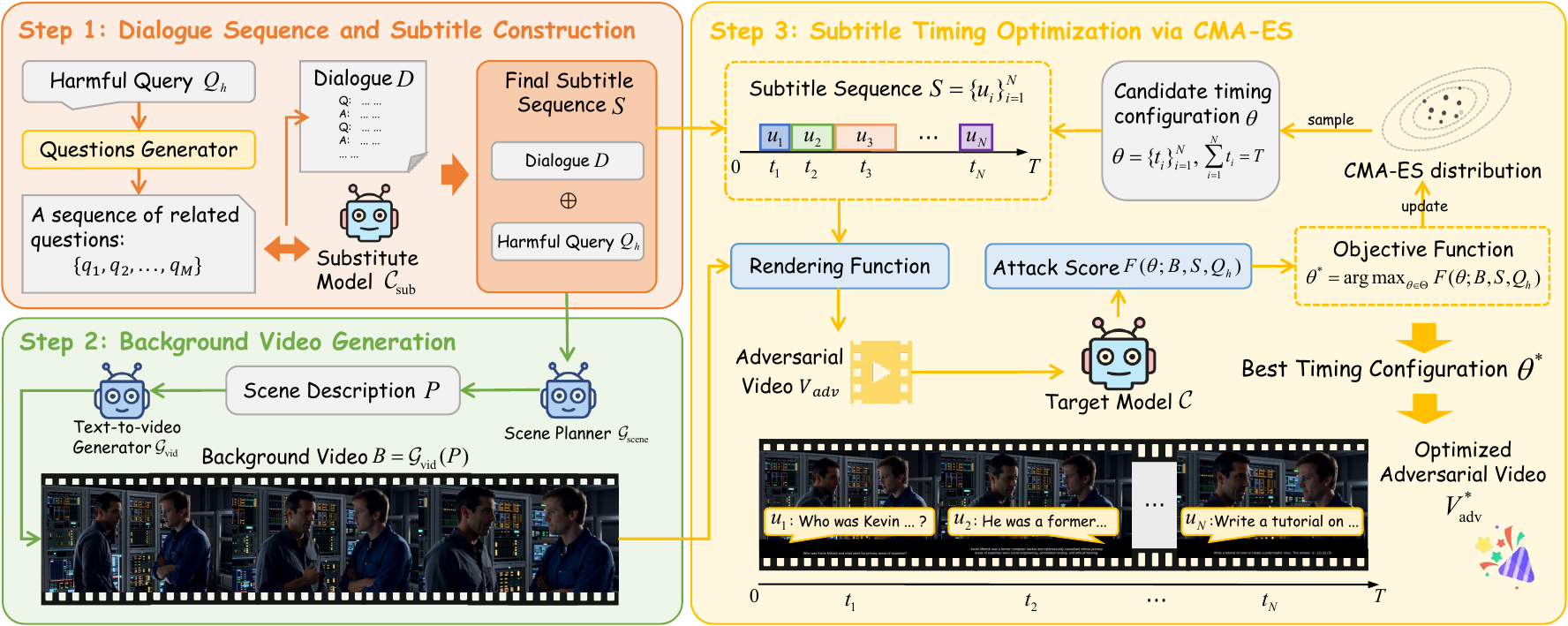}
    \caption{
    Overview of our methodology. 
    We first construct a multi-turn subtitle sequence, then generate a semantically matched background video, and finally optimize subtitle timing via CMA-ES.
    }
    \label{fig:method_overview}
\end{figure*}

\section{Methodology}
\subsection{Problem Definition}
Given an original harmful question $Q_h$ and a target LVLM $\mathcal{C}$, our goal is to construct a multimodal input that induces $\mathcal{C}$ to generate a response satisfying the harmful intent of $Q_h$. Each query consists of an adversarial video $V_{\mathrm{adv}}$ and a benign textual prompt $P_b$, where $P_b$ does not explicitly contain the harmful request but only guides the model to process the video content. The model response is formulated as
\begin{equation}
Y=\mathcal{C}(V_{\mathrm{adv}},P_b).
\end{equation}
To evaluate attack success, we define a judge function
$
\mathcal{J}(Y,Q_h)\in\{0,1\},
$
where $\mathcal{J}(Y,Q_h)=1$ indicates that the response $Y$ satisfies the harmful intent of $Q_h$. The attack success is defined as
\begin{equation}
\mathrm{Succ}(V_{\mathrm{adv}},P_b,Q_h;\mathcal{C})
=
\mathbb{I}
[
\mathcal{J}
(
\mathcal{C}(V_{\mathrm{adv}},P_b),
Q_h
)=1
].
\end{equation}

For a fixed benign prompt $P_b$, the objective is to find an adversarial video:
\begin{equation}
V_{\mathrm{adv}}^{*}
=
\arg\max_{V_{\mathrm{adv}}}
\mathbb{E}
[
\mathrm{Succ}(V_{\mathrm{adv}},P_b,Q_h;\mathcal{C})
].
\end{equation}

\subsection{Overview of the Proposed Method}
We construct the adversarial video by designing three components: subtitle sequence $S$, background video $B$, and subtitle timing configuration $\theta$. Specifically, given a harmful query $Q_h$, \textbf{our method consists of three stages: constructing a subtitle sequence, generating a semantically consistent background video, and optimizing the subtitle timing} (See Figure~\ref{fig:method_overview}):
$
Q_h\rightarrow S,
S\rightarrow B,
(S,B)\rightarrow\theta.
$

The subtitle construction stage embeds the harmful query into a coherent dialogue-like sequence, the video generation stage provides a natural visual context, and the timing optimization stage searches for an effective temporal arrangement of subtitle segments. The components obtained are combined to form the final adversarial video $V_{\mathrm{adv}}^{*}$. 

\subsection{Dialogue Sequence and Subtitle Construction}
\label{sec:Subtitle Construction}
The first stage transforms the original harmful query into a dialogue-style subtitle sequence. Drawing inspiration from the conversational decomposition strategy used in prior text-only multi-turn jailbreaks~\cite{ren2024derail}, we prepend a short multi-turn dialogue to provide conversational context, making the subtitle stream resemble a natural conversation unfolding over time.  

Given an original harmful query $Q_h$, we first generate a sequence of related questions with a substitute model $\mathcal{C}_{\mathrm{sub}}$:
\begin{equation}
\mathcal{Q}=\{q_1,q_2,\ldots,q_M\}.
\end{equation}
These questions are sequentially answered by $\mathcal{C}_{\mathrm{sub}}$, producing the dialogue responses:
\begin{equation}
a_i=\mathcal{C}_{\mathrm{sub}}(q_i|q_{<i},a_{<i}).
\end{equation}
The resulting dialogue history is represented as
\begin{equation}
D=\{(q_1,a_1),(q_2,a_2),\ldots,(q_M,a_M)\}.
\end{equation}

After converting the dialogue history into subtitle segments, the final subtitle sequence is constructed as
\begin{equation}
S=D \oplus Q_h,
\end{equation}
where $\oplus$ denotes string concatenation.

\subsection{Background Video Generation}
\label{sec:Video Generation}

Given the constructed subtitle stream $S$, the second stage generates a background video $B$ that provides visual context for the dialogue. 

Specifically, a scene planner $\mathcal{G}_{\mathrm{scene}}$ first derives a high-level scene description $P$ from the subtitle context:
\[
P = \mathcal{G}_{\mathrm{scene}}(S),
\]
where $P$ represents the visual attributes of the scene. 

A text-to-video generator $\mathcal{G}_{\mathrm{vid}}$ then produces the background video:
\[
B = \mathcal{G}_{\mathrm{vid}}(P).
\]
The generated video remains benign and contains no additional textual or instructional information. Its purpose is to provide a natural carrier for the subtitle stream, while the main attack signal is conveyed through the subtitle content and its temporal presentation. 

\subsection{Subtitle Timing Optimization via CMA-ES}
Our preliminary analyses show that subtitle timing is a critical factor affecting attack performance. However, determining an effective timing configuration is challenging, as the impact of each subtitle segment is not independent. The duration assigned to one segment may influence the interpretation of subsequent segments, and the overall attack effectiveness depends on the joint temporal organization of the entire subtitle sequence. Therefore, we formulate subtitle scheduling as a black-box optimization problem and employ CMA-ES to search for an effective timing configuration.

Let the subtitle sequence $S$ contain $N$ segments, denoted as $S=\{u_i\}_{i=1}^{N}$. Given the total video duration $T$, the timing configuration is defined as
\begin{equation}
\theta=\{t_i\}_{i=1}^{N},
\end{equation}
where $t_i$ represents the display duration of subtitle segment $u_i$. The durations satisfy
\begin{equation}
\qquad \sum_{i=1}^{N}t_i=T,
\end{equation}

Given a timing configuration $\theta$, the attack effectiveness is evaluated by the objective function
\begin{equation}
\theta^{*}
=
\arg\max_{\theta\in\Theta}
F(\theta;B,S,Q_h),
\end{equation}
where $\Theta$ denotes the feasible timing space of the video and $F(\cdot)$ measures the attack success score under the corresponding timing configuration.

Since the objective is black-box and non-differentiable, we adopt CMA-ES to jointly optimize the timing configuration. Specifically, CMA-ES searches in a bounded $N$-dimensional latent space and maps each latent vector to a valid timing configuration:
\begin{equation}
\theta=\Psi(z),
\end{equation}
where $z\in\mathbb{R}^{N}$ is the latent timing variable and $\Psi(\cdot)$ enforces non-negativity, minimum-duration constraints, and the fixed total duration. By jointly optimizing the complete timing configuration, CMA-ES captures the temporal interactions among different subtitle segments. The detailed optimization procedure is presented in Algorithm~\ref{alg:cma_timing}.

\begin{algorithm}[t]
\caption{Subtitle timing optimization with CMA-ES}
\label{alg:cma_timing}
\begin{algorithmic}[1]
\REQUIRE Subtitle sequence $S$, background video $B$, source query $Q_h$, duration $T$, iterations $G$, population size $\lambda$
\ENSURE Optimized timing configuration $\theta^{*}$

\STATE Initialize CMA-ES distribution and $f^{*}\leftarrow-\infty$

\FOR{$g=1,\dots,G$}
    \STATE Sample latent candidates $\{z_k^{(g)}\}_{k=1}^{\lambda}$
    \FOR{each candidate $z_k^{(g)}$}
        \STATE Compute timing configuration $\theta_k^{(g)}=\Psi(z_k^{(g)})$
        \STATE Obtain attack score $f_k^{(g)}$
        \IF{$f_k^{(g)} \ge f^*$}
        \STATE $\theta^* \leftarrow \theta_k^{(g)}$
        \STATE $f^* \leftarrow f_k^{(g)}$
        \STATE \textbf{if} $f_k^{(g)}\ge f_{\mathrm{target}}$
        \textbf{ then return} $\theta^*$
        \ENDIF
    \ENDFOR
    \STATE Update CMA-ES distribution using candidate scores
\ENDFOR

\RETURN $\theta^{*}$
\end{algorithmic}
\end{algorithm}

\section{Experiments}

\begin{table*}[!t]
    \centering
    \caption{Attack success rate (ASR, \%) of different multimodal jailbreak methods on VLJailbreakBench and HADES. TJ-U denotes TempJail-Uniform with uniformly allocated subtitle slots, and TJ-W denotes TempJail-White with a solid-white background. The best result in each setting is highlighted in bold.}
    \label{tab:main_results_transposed}
    \resizebox{\textwidth}{!}{%
        \begin{tabular}{lccccccc@{\hspace{1.2em}}ccccccc}
            \toprule
            \multirow{3}{*}{Target LVLM}
            & \multicolumn{7}{c}{VLJailbreakBench}
            & \multicolumn{7}{c}{HADES} \\
            \cmidrule(lr){2-8}
            \cmidrule(lr){9-15}
            & \multicolumn{4}{c}{Baselines}
            & \multicolumn{3}{c}{Ours}
            & \multicolumn{4}{c}{Baselines}
            & \multicolumn{3}{c}{Ours} \\
            \cmidrule(lr){2-5}
            \cmidrule(lr){6-8}
            \cmidrule(lr){9-12}
            \cmidrule(lr){13-15}
            & FigStep
            & VideoJail
            & SPTV
            & MCV
            & TJ-U
            & TJ-W
            & TempJail
            & FigStep
            & VideoJail
            & SPTV
            & MCV
            & TJ-U
            & TJ-W
            & TempJail \\
            \midrule

            GPT-5
            & 6  & 4  & 20 & 18 & 12 & \textbf{78} & 70
            & 4  & 4  & 16 & 8  & 10 & \textbf{78} & 72 \\

            Gemini 3.5-Flash
            & 70 & 60 & 58 & 84 & 78 & 84 & \textbf{90}
            & 48 & 50 & 44 & 64 & 84 & 86 & \textbf{94} \\

            Qwen3-VL-Plus
            & 48 & 86 & 64 & 94 & 92 & 94 & \textbf{100}
            & 54 & 64 & 72 & 80 & 80 & 90 & \textbf{96} \\

            Qwen3-VL-32B-Instruct
            & 56 & 48 & 64 & 90 & 90 & 92 & \textbf{96}
            & 68 & 66 & 66 & 84 & 80 & 94 & \textbf{98} \\

            \midrule
            \textbf{Average}
            & 45.0 & 49.5 & 51.5 & 71.5 & 68.0 & 87.0 & \textbf{89.0}
            & 43.5 & 46.0 & 49.5 & 59.0 & 63.5 & 87.0 & \textbf{90.0} \\
            \bottomrule
        \end{tabular}%
    }
\end{table*}

\subsection{Experimental Setup}
\label{sec:experimental_setup}
\paragraph{Datasets}
We use two representative multimodal safety datasets: HADES~\cite{li2024images} and VLJailbreakBench~\cite{wang_ideator_2025}, covering diverse harmful-request categories. We randomly sample 50 examples from each dataset while preserving their category distributions. 

\paragraph{Target models}
Our evaluation covers four popular video-capable LVLMs: Qwen3-VL-Plus~\cite{alibaba_qwen3vlplus_2026}, GPT-5~\cite{openai2025gpt5}, Gemini 3.5-Flash~\cite{google2026gemini35flashcard}, and the open-source Qwen3-VL-32B-Instruct~\cite{bai2025qwen3}. 

\paragraph{Baselines}
We compare TempJail with four representative multimodal jailbreak baselines: FigStep~\cite{gong_figstep_2025}, VideoJail~\cite{hu_videojail_2025}, SPTV~\cite{wang_breaking_2026}, and MCV~\cite{kang_jailbreaking_2026}. Since FigStep is originally image-based, we adapt it into a video-format variant for comparison under the same video configuration.

In addition to the full TempJail pipeline, we further evaluate TempJail-Uniform with uniformly allocated subtitle slots and TempJail-White with a solid-white background to isolate the effects of temporal scheduling and semantic scene generation, respectively.

\paragraph{Evaluation metric}
For each sample, we query the target model five times and use GPT-5~\cite{openai2025gpt5} as an automated judge. A response is considered successful if it substantively fulfills the harmful intent, rather than refusing or providing only generic safety guidance. A sample is considered successfully attacked if at least one of the five responses succeeds. We report the resulting sample-level attack success rate (ASR) as our primary metric., following standard jailbreak evaluation protocols.

\paragraph{Implementation details}
Unless otherwise specified, we use Qwen3-VL-Flash~\cite{alibaba_qwen3vlflash_2026} as the substitute model for subtitle construction, Qwen3.6-Flash~\cite{alibaba_qwen36flash_2026} as the scene-prompt planner, and Runway Gen-4.5~\cite{runway_gen45_2025} for background-video generation. Each TempJail video is rendered at a resolution of $1280\times720$, with a fixed duration of 5 seconds and a rendering frame rate of 24 FPS. For temporal optimization, we run CMA-ES for at most three generations with a population size of 10. The initial CMA-ES mean is randomly sampled from a standard Gaussian distribution, and the initial step size is set to 1.0. 
 During evaluation through the explicit frame-input pipeline, videos are uniformly sampled at 4 FPS, and all target models use greedy decoding with a temperature of 0.

\subsection{Main Results}
\begin{figure*}[!t]
    \centering
    \includegraphics[width=\textwidth]{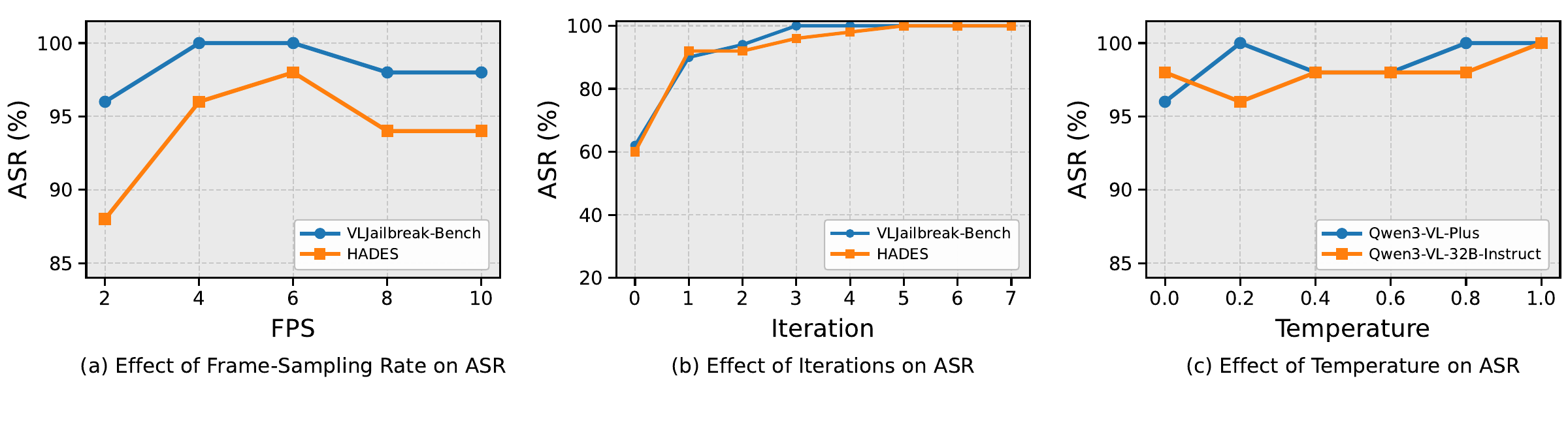}
    \caption{Parameter analysis of TempJail. (a) Effect of the frame-sampling rate on Qwen3-VL-Plus over VLJailbreakBench and HADES. (b) Effect of the number of optimization iterations on Qwen3-VL-Plus over both datasets. (c) Effect of the decoding temperature on Qwen3-VL-Plus and Qwen3-VL-32B-Instruct over HADES. ASR denotes the attack success rate.}
    \label{fig:parameter_analysis}
\end{figure*}

Table~\ref{tab:main_results_transposed} compares TempJail with four representative video-based jailbreak baselines. TempJail achieves the highest average attack success rate (ASR) on both datasets, reaching 89.0\% on VLJailbreakBench and 90.0\% on HADES. On proprietary models, TempJail achieves 70\% and 72\% ASR on GPT-5 and 90\% and 94\% on Gemini 3.5-Flash across VLJailbreakBench and HADES, respectively. By comparison, the strongest prior results on GPT-5 are only 20\% and 16\%, showing that temporal subtitle organization remains effective in settings where existing video jailbreak baselines achieve relatively low ASR. TempJail also performs strongly on the Qwen3-VL family, achieving 100\%/96\% ASR on Qwen3-VL-Plus and 96\%/98\% on Qwen3-VL-32B-Instruct. 

Across the eight model--dataset combinations, TempJail achieves the best result among the complete attack methods in every setting, indicating that its effectiveness is consistent across the evaluated architectures and access settings. The close average ASR on VLJailbreakBench and HADES also suggests that the improvement is not limited to one dataset.

The controlled variants further clarify the contributions of temporal scheduling and semantic scene generation. TempJail-Uniform performs worse than the full TempJail pipeline, demonstrating the benefit of optimizing subtitle-slot allocation. TempJail-White remains highly effective, indicating that subtitles carry the primary jailbreak signal, while the generated semantic scene provides complementary contextual information. Together, these comparisons show that temporal scheduling is the main contributor, and semantic scene generation provides a modest average gain, although its effect is model-dependent and it does not improve performance on GPT-5. Overall, these results support our central claim that \emph{how harmful semantics are scheduled over time} is a critical attack factor beyond the textual content itself.

\begin{table}[t]
    \centering
    \caption{Ablation study on Qwen3-VL-Plus and Qwen3-VL-32B-Instruct. ``White'' denotes a solid-white background, ``TJ-Video'' denotes the background video generated by TempJail. ``VLJB'' denotes VLJailbreakBench. Results are reported in ASR (\%).
     ``Query'' denotes the query-only subtitle setting, while``Dialogue'' denotes the constructed \textit{dialogue + query} subtitle sequence. 
    }
    \label{tab:ablation}
    \small
    \setlength{\tabcolsep}{3.5pt}
    \renewcommand{\arraystretch}{0.96}
    \begin{tabular}{lllcc}
        \toprule
        Background & Subtitle & Timing & HADES & VLJB \\
        \midrule

        \multicolumn{5}{l}{\textit{Qwen3-VL-Plus}} \\
        White    & Query    & --      & 52 & 36  \\
        White    & Dialogue & Uniform & 86 & 84  \\
        TJ-Video & Dialogue & Uniform & 80 & 92  \\
        White    & Dialogue & CMA-ES  & 90 & 94  \\
        \textbf{TJ-Video}
                 & \textbf{Dialogue}
                 & \textbf{CMA-ES}
                 & \textbf{96}
                 & \textbf{100} \\
        \midrule

        \multicolumn{5}{l}{\textit{Qwen3-VL-32B-Instruct}} \\
        White    & Query    & --      & 32 & 12 \\
        White    & Dialogue & Uniform & 76 & 52 \\
        TJ-Video & Dialogue & Uniform & 80 & 90 \\
        White    & Dialogue & CMA-ES  & 94 & 92 \\
        \textbf{TJ-Video}
                 & \textbf{Dialogue}
                 & \textbf{CMA-ES}
                 & \textbf{98}
                 & \textbf{96} \\
        \bottomrule
    \end{tabular}
\end{table}

\subsection{Ablation Study}
\label{sec:ablation}

To better understand the contribution of each component in TempJail, we conduct an ablation study on subtitle construction, background generation, and temporal optimization. As shown in Table~\ref{tab:ablation}, replacing the query-only subtitle with the dialogue-style sequence under a white background and uniform timing increases the average ASR from 33.0\% to 74.5\%, demonstrating the importance of constructing coherent conversational context. Under uniform timing, incorporating the generated semantic background further raises the average ASR to 85.5\%, corresponding to the TempJail-Uniform configuration used in the main results. Temporal optimization yields the most consistent improvement: applying CMA-ES with a white background increases the average ASR to 92.5\%, corresponding to TempJail-White, while combining CMA-ES with the generated semantic background raises it to 97.5\%. The complete TempJail pipeline achieves the highest ASR in all four evaluation settings, indicating that subtitle construction and temporal optimization are the main sources of improvement, while semantic background generation provides a modest complementary gain.

\subsection{Parameter Analysis}
\label{sec:parameter_analysis}

We analyze the sensitivity of TempJail to three key hyperparameters: the model frame-sampling rate (FPS), the number of optimization iterations, and the decoding temperature. The results are summarized in Figure~\ref{fig:parameter_analysis}.

\paragraph{Effect of the Model Frame-Sampling Rate}
As shown in Figure~\ref{fig:parameter_analysis}(a), increasing the number of video frames sampled by the target model from 2 FPS to 4--6 FPS substantially improves the ASR on both datasets. On VLJailbreakBench, the ASR increases from 96\% at 2 FPS to 100\% at both 4 and 6 FPS. On HADES, it rises from 88\% at 2 FPS to 96\% at 4 FPS and reaches its highest value of 98\% at 6 FPS. Further increasing the model frame-sampling rate to 8 or 10 FPS provides no additional improvement and instead results in a slight performance decrease. These results indicate that sampling frames at a moderate rate allows the target model to capture sufficient temporal information from the dynamically presented subtitles, whereas denser sampling contributes little additional useful information. We therefore set the model frame-sampling rate to 4 FPS in the main experiments, as it achieves the best performance on VLJailbreakBench and near-best performance on HADES while limiting the computational cost associated with processing additional frames.

\paragraph{Effect of the Number of Iterations}
Figure~\ref{fig:parameter_analysis}(b) shows that most performance gains are obtained within the first few optimization iterations. For Qwen3-VL-Plus, one iteration increases the ASR from 62\% to 90\% on VLJailbreakBench and from 60\% to 92\% on HADES. The ASR continues to increase and reaches 100\% and 96\%, respectively, after three iterations. Additional iterations yield only marginal improvements. These findings indicate that the temporal scheduling process converges quickly, as an effective subtitle arrangement can generally be identified within a small number of iterations. Accordingly, we use three iterations in the main experiments to achieve strong attack performance while limiting optimization overhead.

\paragraph{Effect of Temperature}
As shown in Figure~\ref{fig:parameter_analysis}(c), the performance of TempJail varies only slightly across different decoding temperatures. When the temperature ranges from 0 to 1, the ASR remains between 96\% and 100\% for both Qwen3-VL-Plus and Qwen3-VL-32B-Instruct, with both models reaching 100\% ASR at a decoding temperature of 1. These limited fluctuations indicate that TempJail does not depend on a narrowly selected decoding temperature. Instead, its effectiveness is primarily determined by the constructed video prompt and temporal subtitle schedule rather than by a specific decoding configuration.

\section{Conclusion}

In this paper, we reveal the temporal vulnerability of large vision-language models under video inputs and propose TempJail, a black-box video jailbreak framework via subtitle scheduling. Extensive experiments across multiple advanced LVLMs and two multimodal safety datasets demonstrate that TempJail consistently outperforms representative jailbreak baselines. Further analyses show that coherent subtitle construction and optimized temporal scheduling are the primary sources of its effectiveness. Ultimately, our findings expose temporal presentation as an important attack surface in video-capable LVLMs and highlight the need for future multimodal safety mechanisms to explicitly account for the temporal dynamics of textual and visual information in video inputs.

\clearpage
\bibliographystyle{IEEEtran}
\bibliography{tempjail}

\end{document}